\documentclass[letterpaper]{article} 
\usepackage[preprint]{aaai2027}  
\usepackage[hyphens]{url}  
\usepackage{graphicx} 
\usepackage{natbib}  
\usepackage{caption} 
\usepackage{amsmath}
\usepackage{amssymb}
\usepackage{booktabs}
\usepackage{algorithm}
\usepackage{algpseudocode}
\usepackage{tikz}
\usetikzlibrary{arrows.meta}

\newcommand{\MHP}{\mathcal{M}_{\mathrm{HP}}}
\newcommand{\MC}{\mathcal{M}_{\mathcal{C}}}
\newcommand{\MBF}{\mathcal{M}_{\mathrm{BF16}}}
\newcommand{\MFP}{\mathcal{M}_{\mathrm{FP4}}}
\newcommand{\LQAT}{\mathcal{L}_{\mathrm{QAT}}}
\newcommand{\LQAH}{\mathcal{L}_{\mathrm{QAH}}}

\title{Quantization-Aware Healing: A Practical Recipe for Recovering Compressed, 4-Bit LLMs}

\author{
    Bakbergen Ryskulov,
    Iker Garc\'ia-Ferrero,
    David Montero,
    David Jansen,
    Ali Hashemi,\\
    Jezabel R. Garcia,
    Antonio Tiene,
    Rom\'an Or\'us
}
\affiliations{
    Multiverse Computing\\
    \{bakbergen.ryskulov, iker.garcia, david.montero, david.jansen, ali.hashemi,\\
    jezabel.garcia, antonio.tiene, roman.orus\}@multiversecomputing.com
}

\begin{document}
\maketitle

\begin{abstract}
Serving large language models cheaply increasingly means shipping models that are both structurally compressed to a fraction of their parameters and quantized to 4 bits. Together these steps degrade reasoning, mathematics, coding, and long-context behavior enough to require a recovery, or \emph{healing}, stage before deployment. The default recipe, quantization-aware training (QAT), re-fits the compressed, quantized model to hard labels; in our pipeline it converged slowly and collapsed past its peak. We adopted \textbf{Quantization-Aware Healing (QAH)} instead. Because a structurally compressed model is never independently trained at full precision, its bfloat16 checkpoint is a distillation-recovered approximation of the original; QAH distills the 4-bit student directly from the original, uncompressed model. On a GPT-OSS 120B$\to$60B$\to$MXFP4 pipeline, the QAH student matches or beats its bfloat16 source on 7 of 9 benchmarks at roughly $4\times$ less weight memory and half the teacher's parameter count, and is released open-weight as Hypernova-60B. Against a matched QAT baseline it reaches a comparable peak about $7\times$ faster and stays stable under continued training, without hand-tuned early stopping. We also report deployment lessons, including a large, reproducible quality gap between distributed-training backends. Our aim is a recipe deployable without a multi-week hyperparameter search.
\end{abstract}
\begin{links}
    \link{Model (Hypernova-60B)}{https://huggingface.co/MultiverseComputingCAI/Hypernova-60B-2605}
\end{links}
\section{Introduction}
\label{sec:intro}

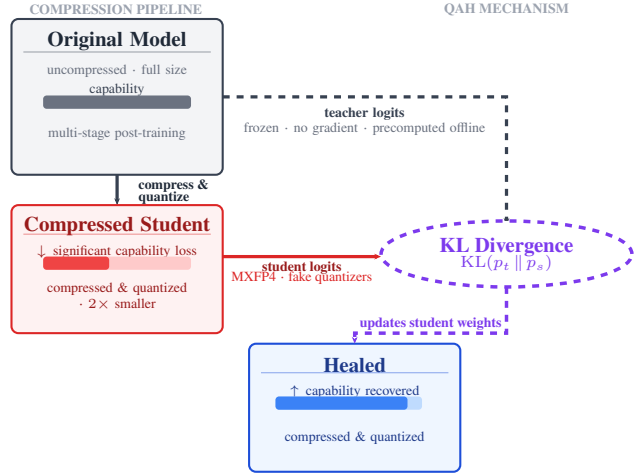
\begin{figure}[t]
  \centering
  \resizebox{0.98\columnwidth}{!}{%
  \begin{tikzpicture}[scale=1]
  \definecolor{origFill}  {RGB}{243, 244, 246}
  \definecolor{origBorder}{RGB}{ 55,  65,  81}
  \definecolor{darkText}  {RGB}{ 31,  41,  55}
  \definecolor{subText}   {RGB}{107, 114, 128}
  \definecolor{barGrayBg} {RGB}{209, 213, 219}
  \definecolor{barGray}   {RGB}{107, 114, 128}
  \definecolor{compFill}  {RGB}{254, 242, 242}
  \definecolor{compBorder}{RGB}{220,  38,  38}
  \definecolor{compText}  {RGB}{153,  27,  27}
  \definecolor{barRedBg}  {RGB}{254, 202, 202}
  \definecolor{barRed}    {RGB}{239,  68,  68}
  \definecolor{healFill}  {RGB}{239, 246, 255}
  \definecolor{healBorder}{RGB}{ 29,  78, 216}
  \definecolor{healText}  {RGB}{ 30,  58, 138}
  \definecolor{barBlueBg} {RGB}{191, 219, 254}
  \definecolor{barBlue}   {RGB}{ 59, 130, 246}
  \definecolor{qahPurple} {RGB}{124,  58, 237}
  \definecolor{headerGray}{RGB}{156, 163, 175}
  \node[font=\tiny\bfseries, text=headerGray] at (1.68, 7.10) {COMPRESSION PIPELINE};
  \node[font=\tiny\bfseries, text=headerGray] at (7.60, 7.10) {QAH MECHANISM};
  \draw[fill=origFill, draw=origBorder, line width=0.9pt, rounded corners=3pt] (0.10, 4.60) rectangle (3.30, 6.96);
  \node[font=\small\bfseries, text=darkText, text width=3.00cm, align=center] at (1.70, 6.62) {Original Model};
  \draw[origBorder, opacity=0.5, line width=0.25pt] (0.28, 6.46) -- (3.12, 6.46);
  \node[font=\tiny, text=subText, text width=2.90cm, align=center] at (1.70, 6.18) {uncompressed $\cdot$ full size};
  \node[font=\tiny, text=origBorder, text width=2.90cm, align=center] at (1.70, 5.88) {capability};
  \fill[barGrayBg, rounded corners=1.5pt] (0.58, 5.60) rectangle (2.82, 5.80);
  \fill[barGray,   rounded corners=1.5pt] (0.58, 5.60) rectangle (2.82, 5.80);
  \node[font=\tiny, text=subText, text width=2.90cm, align=center] at (1.70, 5.22) {multi-stage post-training};
  \draw[-{Stealth[length=3pt, width=2.5pt]}, draw=origBorder, line width=1.4pt] (1.70, 4.60) -- (1.70, 4.14);
  \node[font=\tiny\bfseries, text=darkText, anchor=west] at (1.90, 4.44) {compress \&};
  \node[font=\tiny\bfseries, text=darkText, anchor=west] at (1.90, 4.26) {quantize};
  \draw[fill=compFill, draw=compBorder, line width=0.9pt, rounded corners=3pt] (0.10, 2.20) rectangle (3.30, 4.14);
  \node[font=\small\bfseries, text=compText, text width=3.00cm, align=center] at (1.70, 3.82) {Compressed Student};
  \draw[compBorder, opacity=0.5, line width=0.25pt] (0.28, 3.70) -- (3.12, 3.70);
  \node[font=\tiny, text=compText, text width=2.90cm, align=center] at (1.70, 3.44) {$\downarrow$\, significant capability loss};
  \fill[barRedBg, rounded corners=1.5pt] (0.58, 3.16) rectangle (2.82, 3.36);
  \fill[barRed,   rounded corners=1.5pt] (0.58, 3.16) rectangle (1.58, 3.36);
  \node[font=\tiny, text=compText, text width=2.90cm, align=center] at (1.70, 2.76) {compressed \& quantized\\$\cdot$~$2\times$ smaller};
  \draw[fill=healFill, draw=healBorder, line width=0.9pt, rounded corners=3pt] (3.70, 0.10) rectangle (6.90, 2.04);
  \node[font=\small\bfseries, text=healText, text width=3.10cm, align=center] at (5.30, 1.72) {Healed};
  \draw[healBorder, opacity=0.5, line width=0.25pt] (3.88, 1.60) -- (6.72, 1.60);
  \node[font=\tiny, text=healText, text width=2.90cm, align=center] at (5.30, 1.32) {$\uparrow$\, capability recovered};
  \fill[barBlueBg, rounded corners=1.5pt] (4.10, 1.04) rectangle (6.32, 1.24);
  \fill[barBlue,   rounded corners=1.5pt] (4.10, 1.04) rectangle (6.10, 1.24);
  \node[font=\tiny, text=healText, text width=2.90cm, align=center] at (5.30, 0.62) {compressed \& quantized};
  \draw[-{Stealth[length=3pt, width=2.5pt]}, draw=origBorder, line width=1.4pt, dashed] (3.30, 5.78) -- (7.60, 5.78) -- (7.60, 3.84);
  \node[font=\tiny\bfseries, text=origBorder, text width=3.40cm, align=center] at (5.45, 5.52) {teacher logits};
  \node[font=\tiny, text=subText, text width=3.80cm, align=center] at (5.45, 5.30) {frozen $\cdot$ no gradient $\cdot$ precomputed offline};
  \draw[-{Stealth[length=3pt, width=2.5pt]}, draw=compBorder, line width=1.4pt] (3.30, 3.36) -- (5.70, 3.36);
  \node[font=\tiny\bfseries, text=compText] at (4.50, 3.20) {student logits};
  \node[font=\tiny, text=compBorder] at (4.50, 3.04) {MXFP4 $\cdot$ fake quantizers};
  \draw[fill=white, draw=qahPurple, line width=1.5pt, dashed] (7.60, 3.40) ellipse (1.90 and 0.50);
  \node[font=\small\bfseries, text=qahPurple] at (7.60, 3.50) {KL Divergence};
  \node[font=\scriptsize, text=qahPurple] at (7.60, 3.26) {$\mathrm{KL}(p_t \,\|\, p_s)$};
  \draw[-{Stealth[length=3pt, width=2.5pt]}, draw=qahPurple, line width=1.5pt, dashed] (7.60, 2.90) -- (7.60, 2.20) -- (5.30, 2.20) -- (5.30, 2.04);
  \node[font=\tiny\bfseries, text=qahPurple] at (6.45, 2.34) {updates student weights};
  \end{tikzpicture}}
  \caption{%
    \textbf{QAH overview.}
    Structural compression followed by 4-bit quantization sharply reduces
    capability. QAH heals the compressed, quantized student by distilling from
    the \emph{original} uncompressed model (frozen teacher, dashed arrow),
    rather than re-fitting hard labels or distilling from the recovered
    full-precision checkpoint.}
  \label{fig:teaser}
\end{figure}

The memory and compute cost of large language models (LLMs) has made low-precision, reduced-size deployment a practical necessity rather than an option. A now-common production pattern combines two compression steps in sequence: \emph{structural compression}, which reduces a model to a fraction of its original parameter count by modifying its architecture, and \emph{4-bit quantization}, which further cuts memory and inference cost. Applied together, these steps deliver large efficiency gains, but at a measurable cost in reasoning, mathematical problem-solving, code generation, and long-context ability. For models that have already been through multi-stage post-training (supervised fine-tuning, RLHF/RLAIF, model merging, agentic-behavior tuning), this degradation compounds across stages and must be recovered before the model can be shipped. We refer to this recovery step as \emph{healing}.

The default healing recipe is \emph{quantization-aware training} (QAT)~\citep{jacob2018quantization,liu2023llmqat}, which inserts fake-quantizers into the forward pass and continues training under a task (cross-entropy) loss. In our deployment setting we found QAT to be both expensive and operationally fragile: it converges slowly, and if training is allowed to continue past its peak it collapses, so shipping a QAT checkpoint safely requires careful early stopping against a held-out signal. Recent work on 4-bit inference reports the same fragility and proposes \emph{quantization-aware distillation} (QAD), which replaces the task loss with a KL objective against a frozen full-precision copy of the model~\citep{nvidia2026qad}. QAD is a strong recipe when compression is quantization-only, because a full-precision counterpart of the quantized model exists by construction.

Under \emph{structural} compression that assumption breaks. The natural QAD teacher is the structurally compressed bfloat16 checkpoint that precedes quantization---but that checkpoint is not an independently trained model. It is a distillation-recovered approximation of the original, and it already carries the capacity loss from compression. Distilling the 4-bit student from it anchors the student to a degraded target and caps its accuracy at the recovered checkpoint's ceiling. This left us with a concrete deployment question: \emph{how should we heal a model that has been both structurally compressed and 4-bit quantized?}

Our answer, and the recipe this paper documents, is \textbf{Quantization-Aware Healing (QAH)}. QAH heals the compressed, quantized student by distilling from the \emph{original, uncompressed} model---a strictly stronger teacher---rather than from the recovered checkpoint (Figure~\ref{fig:teaser}). The teacher and student no longer share an architecture; the student is supervised only through the teacher's output distribution, which is architecture-agnostic. Concretely, on a two-stage pipeline that compresses a GPT-OSS 120B model to 60B and then re-quantizes it to MXFP4, the QAH student matches or exceeds its own bfloat16 source on 7 of 9 benchmarks, and reaches the original 120B model on LiveCodeBench---while using roughly $4\times$ less weight memory and running at half the teacher's parameter count. The practical reading is that the quantization stage is not a lossy postprocessing step to be minimized, but a second opportunity to apply teacher supervision that the bfloat16 checkpoint never received. The recipe described here was used to produce \textbf{Hypernova-60B}, an open-weight model released by Multiverse Computing under Apache~2.0; the public release incorporates further training beyond the pipeline evaluated in this paper, so the numbers we report are our own measurements of the pipeline rather than the released checkpoint's published figures.

This is an experience-and-recipe paper rather than a claim of a new algorithmic primitive: QAH combines well-understood ingredients (knowledge distillation and fake-quantized training) in a way that fits the realities of a compress-then-quantize production pipeline. Our contribution is in identifying where the standard recipe fails in that pipeline, in a recipe that works reliably without a hyperparameter search, and in the operational lessons that made it deployable.

\paragraph{Contributions.}
\begin{itemize}
  \item We identify why the standard QAD recipe is a poor fit once structural compression precedes quantization: the only available full-precision teacher is a recovered checkpoint that caps the student's accuracy. We describe \textbf{QAH}, which distills from the original uncompressed model instead (\S\ref{sec:method}).
  \item On a GPT-OSS 120B$\to$60B$\to$MXFP4 pipeline, the QAH student matches or beats its bfloat16 source on 7 of 9 benchmarks at roughly $4\times$ lower weight memory, and reaches the 120B model on LiveCodeBench (\S\ref{sec:results}).
  \item We report the deployment lessons that made QAH usable in practice: it converges about $7\times$ faster than a matched QAT baseline and does not require hand-tuned early stopping, and we document a large, reproducible quality gap between distributed-training backends that practitioners should be aware of (\S\ref{sec:lessons}).
\end{itemize}

\section{Background and Related Work}
\label{sec:related}

\paragraph{Low-precision formats.}
Microscaling formats (MX, MXFP4, MXFP6, NVFP4) replace per-tensor exponent overhead with block-wise scaling factors, giving aggressive memory and compute reductions while preserving most of the bfloat16 forward pass~\citep{rouhani2023microscaling,micikevicius2022fp8}. MXFP4 quantizes weights only and is the format behind the public GPT-OSS release~\citep{openai2025gptoss}; NVFP4 quantizes weights and activations and is used by Nemotron~3 Nano~\citep{nvidia2026qad}. Post-training quantization (PTQ) methods---GPTQ~\citep{frantar2023gptq}, AWQ~\citep{lin2024awq}, SmoothQuant~\citep{xiao2023smoothquant}---suffice at moderate compression ratios but leave a non-negligible 4-bit accuracy gap on reasoning and coding for models above roughly 10B parameters, which is the regime we operate in.

\paragraph{Quantization-aware training.}
QAT inserts straight-through-estimator fake-quantizers into the forward pass so weights can adapt to the quantized representation~\citep{jacob2018quantization,krishnamoorthi2018quantizing}. For LLMs, LLM-QAT~\citep{liu2023llmqat} closes much of the PTQ gap for moderate-scale dense models, and LR-QAT~\citep{bondarenko2024llmqat} reduces its memory overhead. All of these optimize a task loss, which forces the student to re-acquire behavior already present in the original model---a re-traversal that is doubly costly when the student has also been structurally modified. We use QAT as our baseline (\S\ref{sec:method:qat}).

\paragraph{Knowledge distillation.}
Distillation~\citep{hinton2015distilling} trains a student to match a teacher's output distribution rather than rediscover it from labels; for LLMs, MiniLLM~\citep{gu2024minillm} and DistiLLM~\citep{ko2024distillm} show it outperforms supervised fine-tuning on reasoning and instruction following. Crucially, distillation does not require teacher and student to share an architecture---cross-architecture distillation is standard, from BERT compression (DistilBERT~\citep{sanh2019distilbert}, TinyBERT~\citep{jiao2020tinybert}) to vision-language pruning~\citep{wang2022efficientvlm}. QAH relies on exactly this property, since its teacher (the uncompressed model) and student (the compressed, quantized model) differ in architecture. The idea of combining distillation with quantization itself is not new~\citep{polino2018model}; our contribution is the choice of teacher in the compress-then-quantize setting.

\paragraph{Structural compression.}
A complementary family of methods reduces parameter count by modifying architecture: structured pruning of heads, neurons, or layers~\citep{ma2023llmpruner,xia2022structured}, outlier-aware sparsity~\citep{yin2023outlier,mcgowan2024fishleg}, embedding-dimension slicing (SliceGPT~\citep{ashkboos2024slicegpt}), low-rank decomposition (SVD-LLM~\citep{wang2024svdllm}, ASVD~\citep{yuan2023asvd}), and tensor-network methods~\citep{novikov2015tensorizing,xu2023tensorgpt,tomut2025compactifai}. The common property is that the compressed model is never independently trained at full precision: its weights are derived from the original via a compression operator, and any full-precision version of it is itself a distillation-recovered approximation. This is precisely why a QAD teacher taken from that checkpoint is suboptimal, and why healing under structural compression calls for a different teacher---the original pre-compression model.

\paragraph{Quantization-aware distillation.}
QAD, formalized for NVFP4 by \citet{nvidia2026qad}, starts from the original bfloat16 model as a frozen teacher and trains a fake-quantized student to match its logits via KL divergence. The reported advantages over QAT are no re-traversal of post-training, robustness to data coverage, and single-stage recovery to near-full-precision accuracy. Standard QAD presupposes architectural identity between teacher and student, which holds when compression is quantization-only but not when structural compression comes first. QAH targets that compound-compression regime by distilling directly from the original model, making no assumption about shared layer structure. To our knowledge, healing for models that have undergone \emph{both} structural compression and low-precision quantization has not been studied.

\section{Method}
\label{sec:method}

\subsection{The Compound-Compression Regime}
\label{sec:method:regime}

Let $\MHP$ be the original, uncompressed model, which has already completed multi-stage post-training (SFT, RLHF/RLAIF, and any downstream behavior tuning). We call the pipeline of structural compression followed by low-precision quantization the \emph{compound-compression} regime. In it, $\MHP$ undergoes three modifications before deployment.

\paragraph{Structural compression.}
A compression operator $\mathcal{S}$ reduces the parameter count of $\MHP$ by modifying its architecture---removing attention heads, feed-forward neurons, or transformer layers, or reducing embedding dimensionality. The output is a compressed intermediate $\MC = \mathcal{S}(\MHP)$, $2\times$ smaller in our experiments. \emph{$\MC$ is never trained at full precision:} its weights are derived from $\MHP$ by the operator, and no full-precision checkpoint of the compressed architecture is trained from scratch.

\paragraph{Full-precision recovery.}
Before quantization, $\MC$ is recovered in bfloat16 to a checkpoint $\MBF$ by distilling from $\MHP$ via KL divergence. This adapts the compressed weights to recover capability lost to compression. It is not independent training: the recovered checkpoint is tethered throughout to $\MHP$'s output distribution and inherits the capacity ceiling of the compressed architecture.

\paragraph{Quantization.}
A quantization operator $\mathcal{Q}$ casts $\MBF$ to MXFP4, yielding $\MFP = \mathcal{Q}(\MBF)$. The full pipeline $\MFP = \mathcal{Q}(R(\mathcal{S}(\MHP)))$, with $R$ the recovery operator, defines the regime. Quantization adds a further accuracy gap on top of the compression loss.

\paragraph{Healing goal.}
Healing returns $\MFP' = H(\MFP)$ such that (i) $\MFP'$ approaches $\MHP$ on a target evaluation suite, (ii) $\MFP'$ remains a valid MXFP4 deployment artifact, and (iii) healing does not regress a broader sanity suite. The defining feature of healing, as opposed to generic fine-tuning, is that $\MHP$ exists and is accessible: the target behavior is observable on a frozen teacher, so the student need not rediscover it from labels.

\subsection{Quantization-Aware Healing}
\label{sec:method:qah}

QAH replaces the recovered checkpoint $\MBF$ as teacher with the original model $\MHP$. Teacher and student now have different architectures: $\MHP$ is full-size and uncompressed, while the student $\MFP$ is structurally compressed. Notably, the advantage comes from the teacher being \emph{uncompressed}, not from a difference in precision: $\MHP$ is the original GPT-OSS release, whose weights are already largely MXFP4 (\S\ref{sec:setup}), so QAH distills a compressed 4-bit student from an uncompressed teacher of comparable precision rather than from a higher-precision one. The student is supervised only through the teacher's output distribution. The QAH loss is a teacher--student KL divergence on output logits:
\begin{align}
  \LQAH &= \mathbb{E}_{x \sim \mathcal{D}}\Bigl[
      \mathrm{KL}\Bigl(
        \mathrm{softmax}\!\bigl(\MHP(x)\,/\,\tau\bigr)
        \nonumber\\
  &\hspace{2.5em}\Big\|\;
        \mathrm{softmax}\!\bigl(\MFP(x;\,Q_\theta)\,/\,\tau\bigr)
      \Bigr)
    \Bigr],
  \label{eq:qah}
\end{align}
where $Q_\theta$ are straight-through-estimator MXFP4 fake-quantizers inserted into the student's forward pass, $\tau$ is the distillation temperature ($\tau=1$ throughout), and $\mathcal{D}$ is a healing corpus. The KL uses the standard next-token shift and is averaged over masked positions. Following \citet{nvidia2026qad}, teacher logits are precomputed offline from the released (MXFP4) Hugging Face checkpoint of $\MHP$, once per example, and truncated to the top-$k$ logits ($k=100$) to bound storage. The student sees no hard labels---only the teacher's distribution.

\subsection{Quantization-Aware Training Baseline}
\label{sec:method:qat}

QAT inserts the same fake-quantizers $Q_\theta$ but optimizes a task loss:
\begin{equation}
  \LQAT = \mathbb{E}_{(x,y) \sim \mathcal{D}}\bigl[
      \mathrm{CE}\!\bigl(y,\;\MFP(x;\,Q_\theta)\bigr)
    \bigr],
  \label{eq:qat}
\end{equation}
with $(x,y)$ input--label pairs and $\mathrm{CE}$ next-token cross-entropy. Unlike QAH, QAT does not condition on $\MHP$ at training time; it must re-acquire, from labels, behavior that $\MHP$ already encodes.

\subsection{Making QAH Fit the Memory Budget}
\label{sec:method:system}

Healing at 16k--32k context within a fixed GPU-memory envelope is the main engineering obstacle. A direct implementation of Eq.~\ref{eq:qah} materializes the full $[B,L,V]$ student log-softmax and its autograd graph, whose peak memory $O(BLV)$ becomes prohibitive at $L=32$k with a vocabulary $V$ of order $2{\times}10^5$. We instead adopt the offline top-$K$ logits and fused chunked-KL implementation of \citet{ryskulov2026chunkedkl}: the teacher's top-$k$ logits are cached once rather than recomputed each step, and the KL loss and its gradient are accumulated block-by-block along the sequence, so peak intermediate memory becomes linear in sequence length rather than proportional to vocabulary size while remaining bit-identical to the dense loss. This is what lets QAH train at 32k context on the same hardware as short-context QAT; we refer the reader to that work for the derivation and kernel details. We additionally apply a two-phase masking schedule---an attention-mask-only stability window, then assistant-focused masking for the bulk of training---which further reduces effective sequence length; loss-mask-only training consistently underperformed.

\section{Experimental Setup}
\label{sec:setup}

\paragraph{Models and pipeline.}
We evaluate GPT-OSS 120B and 20B~\citep{openai2025gptoss}, both released with their MoE weights (the large majority of parameters) in MXFP4 and remaining tensors in BF16; we refer to these released checkpoints as MXFP4 throughout. Our pipeline has two stages. First we \emph{compress} each model---120B to 60B and 20B to 9B---using a tensor-network compression operator~\citep{tomut2024compactifai,bercovich2024puzzle,jansen2026blockremoval}, then \emph{recover} each compressed checkpoint by continued bfloat16 training with KL distillation from the corresponding uncompressed GPT-OSS teacher. Second, we \emph{re-quantize} the recovered 60B and 9B checkpoints to MXFP4, again distilling from the uncompressed teacher (QAH) or, for the baseline, training under cross-entropy (QAT).

\paragraph{Data.}
Both stages train on a mixture of NVIDIA Nemotron~\citep{nvidia_nemotron_nano_v3_2025,Nemotron_Cascade_2} and SmolTalk~2~\citep{bakouch2025smollm3} data, covering general-domain, science, coding, mathematics, safety, and reasoning tasks.

\paragraph{Training.}
All runs use 8 nodes of NVIDIA H200 GPUs with FSDP2 for parameter sharding. Both stages optimize a pure KL objective against the uncompressed teacher (QAT excepted), with assistant-focused loss masking. For the MXFP4 QAH stage we use sequence length 32k, global batch size 64, learning rate $5{\times}10^{-6}$, and 400 steps. Quantization-sensitive submodules (embeddings, layer norms, selected attention components) are frozen to limit destructive drift.

\paragraph{Evaluation.}
We evaluate on nine benchmarks: MMLU-Pro~\citep{wang2024mmlupro} (general knowledge), GPQA Diamond~\citep{rein2023gpqa} (science), AIME~2025~\citep{aime25} (mathematics), IFBench~\citep{pyatkin2026generalizing} (instruction following), SciCode~\citep{tian2024scicode} (scientific coding), LiveCodeBench~\citep{jain2024livecodebench} (coding), $\tau^2$-bench~\citep{barres2025tau} (agentic tool use), Aider~\citep{aider} (agentic coding), and AA-LCR~\citep{artificialanalysis2025lcr} (long-context reasoning).

\section{Results}
\label{sec:results}

\begin{figure*}[!htb]
  \centering
  \includegraphics[width=0.85\textwidth]{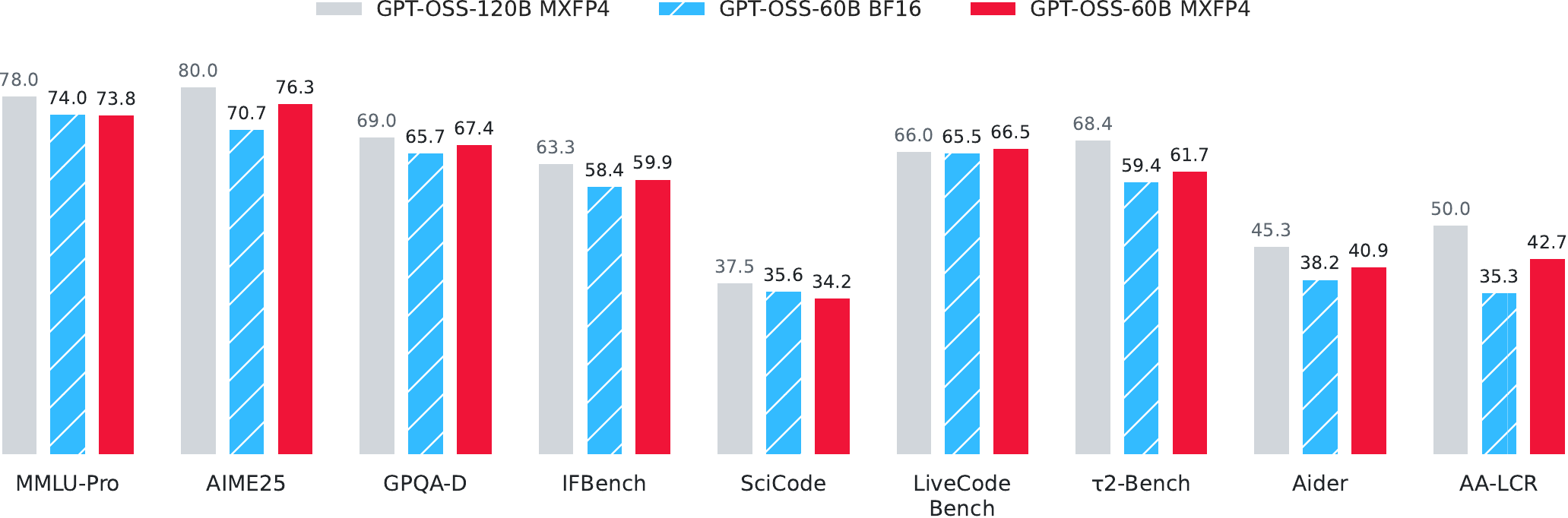}
  \caption{Benchmark performance of three checkpoints in our pipeline: the original \textsc{gpt-oss-120B} (grey), the compressed-and-recovered \textsc{gpt-oss-60B} in bfloat16 (hatched blue), and the same 60B model re-quantized to MXFP4 under QAH (red). The 120B model is the frozen teacher for both the recovery and the quantization stages. The 4-bit QAH student matches or beats its own bfloat16 source on 7 of 9 benchmarks.}
  \label{fig:main_results}
\end{figure*}

\subsection{Quantization as a Second Training Opportunity}
\label{sec:results:main}

Figure~\ref{fig:main_results} compares three checkpoints across nine benchmarks: the original \textsc{gpt-oss-120B} teacher, the 60B bfloat16 student recovered from compression, and the same 60B student re-quantized to MXFP4 under QAH. Naively quantizing the recovered bfloat16 checkpoint to MXFP4 without healing incurs a substantial accuracy drop that QAH must recover; as the figure shows, it more than recovers it on most benchmarks.

The central practical finding is that the 4-bit checkpoint is not a degraded copy of the 16-bit one---it is generally better. The MXFP4 60B student matches or beats its bfloat16 source on 7 of 9 benchmarks; the two exceptions, MMLU-Pro and SciCode, lose by only $0.2$ and $1.4$ points. Gains on the rest are substantial: $+7.4$ on AA-LCR ($42.7$ vs.\ $35.3$), $+5.6$ on AIME~2025 ($76.3$ vs.\ $70.7$), $+2.7$ on Aider ($40.9$ vs.\ $38.2$), and $+2.3$ on $\tau^2$-bench ($61.7$ vs.\ $59.4$).

This is a direct consequence of the QAH recipe rather than of quantization per se: the quantization stage performs a \emph{second} pass of KL distillation against the original 120B distribution, so the quantized student receives teacher supervision that the bfloat16 checkpoint never saw. The deployment-relevant takeaway is that in a distillation-based healing pipeline, the quantization step is not a lossy postprocessing step to be minimized but a place to add capability. We are deliberate about what this does and does not show: it does not establish that 4-bit representations are inherently superior; it shows that, given the extra distillation pass QAH performs at quantization time, the deployable 4-bit artifact is at least as capable as its bfloat16 source at a fraction of the memory (\S\ref{sec:limitations}).

Despite being half the teacher's parameter count and quantized to 4 bits, the QAH student retains most of the teacher's capability. On LiveCodeBench it slightly exceeds the 120B model ($66.5$ vs.\ $66.0$; a difference well within run-to-run noise, so we read it as \emph{matching} the teacher), and on GPQA Diamond it closes to within $1.6$ points ($67.4$ vs.\ $69.0$). MMLU-Pro and IFBench show gaps of $4.2$ and $3.4$ points. The largest residual gap is on AA-LCR ($-7.3$), an extreme long-context benchmark where capacity lost to compression is hardest to recover.

\paragraph{Efficiency.}
These accuracy results come with real serving wins. At 4-bit precision the QAH model uses roughly $4\times$ less weight memory than the bfloat16 student, and at half the teacher's parameter count it roughly halves compute per token, enabling deployment on substantially smaller hardware. For model families released in bfloat16 rather than 4-bit (e.g., Qwen~\citep{qwen3.5}), the combined parameter and precision reduction would amount to roughly $8\times$ less compute per token. QAH therefore inverts the usual low-bit trade-off: instead of trading accuracy for efficiency, it delivers a model that is at once cheaper to serve, lighter in memory, and at least as strong as its bfloat16 counterpart.

\subsection{QAH vs.\ QAT: Cost and Stability}
\label{sec:results:qat}

\begin{figure*}[ht]
  \centering
  \includegraphics[width=0.85\textwidth]{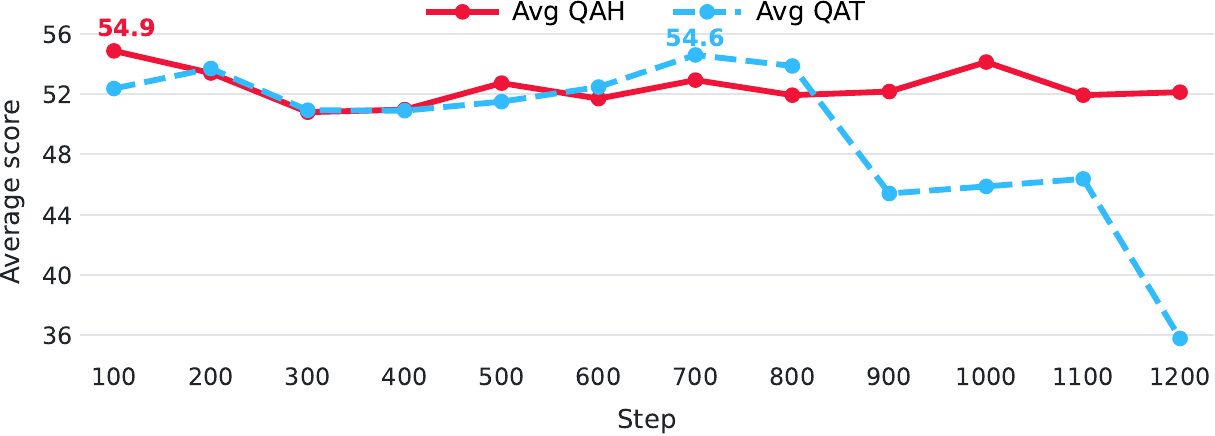}
  \caption{Average performance of QAH and QAT on MMLU-Pro, LiveCodeBench, and GPQA Diamond as training progresses, quantizing GPT-OSS 9B to MXFP4. QAH peaks at $54.9$ in roughly $100$ steps and stays stable through $1200$ steps. QAT reaches a comparable peak ($54.6$) only at step $700$, then collapses, losing nearly $19$ points by step $1200$.}
  \label{fig:qah_vs_qat}
\end{figure*}

For a practitioner, the choice between healing recipes turns less on peak accuracy than on how much it costs to reach that peak and how safe the resulting checkpoint is to ship. Figure~\ref{fig:qah_vs_qat} compares QAH and QAT when quantizing GPT-OSS 9B to MXFP4, plotting average performance on MMLU-Pro, LiveCodeBench, and GPQA Diamond against training steps. This 20B$\to$9B configuration was produced and validated for an internal client deployment (details withheld), giving us a second end-to-end setting---alongside the released 60B model---in which the QAH recipe was applied in practice.

The two recipes reach a comparable peak ($54.9$ for QAH vs.\ $54.6$ for QAT), but differ sharply on the two axes that matter for deployment. First, \textbf{cost}: QAH peaks in about $100$ steps versus roughly $700$ for QAT---about $7\times$ fewer steps; the same pattern holds at 60B, where QAH peaks in roughly $400$ steps. Second, \textbf{stability}: QAH stays within about two points of its peak through all $1200$ steps, whereas QAT collapses once past its peak, losing nearly $19$ points by step $1200$ (from $54.6$ at step $700$ to roughly $36$). Similar QAT-style instability has been reported in concurrent work on NVFP4 distillation~\citep{nvidia2026qad}.

We attribute the asymmetry to the loss. KL distillation against a frozen teacher pins the student to the teacher's distribution and gives it no incentive to drift once matched; the cross-entropy objective keeps pushing the quantized student toward hard labels and eventually erodes unrelated capabilities inherited from the original model. Operationally this is the difference between a recipe that can be run to convergence and shipped, and one that requires hand-tuned early stopping against a held-out signal to avoid deploying a collapsed model. In a production setting the second is a standing operational liability, and it is a large part of why we adopted QAH.

\section{Deployment Lessons}
\label{sec:lessons}

Beyond the recipe itself, three lessons shaped whether healing worked in practice. We report them because they cost us time to learn and are, we believe, transferable.

\paragraph{The distributed backend is a hyperparameter.}
FSDP2~\citep{zhao2023fsdp} and DeepSpeed ZeRO-3~\citep{rajbhandari2020zero,rasley2020deepspeed} are usually treated as throughput-equivalent, interchangeable backends for sharded data parallelism, and in principle the choice should not affect optimization quality. Empirically we found a large and consistent gap. Table~\ref{tab:framework} summarizes the best QAT run under each backend on the 120B student, drawn from a sweep of eleven configurations spanning context lengths $\{2,4,8\}$k, batch sizes $\{128,256\}$, $\{4,8\}$ nodes, and learning rates from $10^{-7}$ to $5{\times}10^{-4}$ (full sweep in Appendix~\ref{app:sweep}). No DeepSpeed configuration reaches the released 120B MXFP4 baseline on GPQA Diamond, plateauing at $65.15$ across the entire sweep, while FSDP2 reaches $73.74$. The gap is largest on reasoning-heavy GPQA Diamond ($-8.6$ points) and persists on MMLU-Pro and AIME~2025. We suspect an interaction between DeepSpeed's mixed-precision communication path and the MXFP4 weight encoding and leave the precise diagnosis to future work; the immediate, actionable lesson is that for MXFP4 distillation workloads the backend must be treated as a tuned hyperparameter, and we default to FSDP2.

\begin{table}[t]
  \centering
  \small
  \begin{tabular}{@{}lcccc@{}}
    \toprule
    \textbf{Backend} & \textbf{GPQA:D} & \textbf{MMLU-P} & \textbf{AIME25} & \textbf{LCB} \\
    \midrule
    DeepSpeed        & $65.15$          & $69.45$          & $76.67$          & $\mathbf{66.79}$ \\
    \textbf{FSDP2}   & $\mathbf{73.74}$ & $\mathbf{70.95}$ & $\mathbf{80.00}$ & $66.42$          \\
    \midrule
    \textit{Target (120B MXFP4)} & $69.0$ & $78.0$ & $80.0$ & $66.0$ \\
    \bottomrule
  \end{tabular}
  \caption{Best QAT run under each distributed backend on the 120B student. The DeepSpeed row is the strongest of eight configurations (8k context, lr $5{\times}10^{-6}$, bs $256$, 8 nodes, 5000 steps); FSDP2 is the strongest of three (2k context, lr $10^{-5}$, bs $128$, 4 nodes, 2200 steps). Across the full DeepSpeed sweep no configuration exceeds $65.15$ on GPQA Diamond. Bold marks the per-column best among QAT runs.}
  \label{tab:framework}
\end{table}

\paragraph{Freeze quantization-sensitive submodules.}
In our sweep, unfreezing layer norms and embedding projections during the quantization stage---especially at higher learning rates---produced checkpoints worse than the unhealed MXFP4 model. Freezing embeddings, layer norms, and selected attention components was necessary for stable healing under both QAT and QAH.

\paragraph{Long-context healing is memory-bound, but tractable.}
Healing at 16k--32k context is the binding constraint on the memory envelope. The chunked-KL implementation of \S\ref{sec:method:system} is what made 32k-context QAH fit the same hardware as short-context QAT: it reduces peak intermediate memory from $O(BLV)$ to $O(BcV)$ while remaining bit-identical to the dense loss, so the long-context capability that compression damages most can actually be healed rather than skipped.

\section{Limitations and Open Questions}
\label{sec:limitations}

We are explicit about scope, in the spirit of reporting what did not get tested alongside what did.

\textbf{The direct QAD baseline is missing.} Our central argument is that distilling from the original model beats distilling from the recovered bfloat16 checkpoint (standard QAD). We motivate this from the quantization literature---a quantized-and-recovered checkpoint is by construction no stronger than the model it approximates---but we did not run the head-to-head experiment (QAH vs.\ QAD-from-$\MBF$ at matched configuration). It is the single most valuable experiment to add, and until it is run our claim of a recovered-teacher ceiling should be read as well-motivated but not directly measured.

\textbf{Single run, small-$n$ benchmarks.} Every number is a single run with no seed variance or confidence intervals, and some benchmarks are small (AIME~2025 has 30 problems), so individual deltas should be read as indicative rather than significant. Where two systems are within a point or two---most notably the LiveCodeBench comparison to the teacher---we read the result as a match, not a win.

\textbf{One model family, one format, one data mixture.} All experiments use GPT-OSS MoE transformers ($120$B$\to$$60$B and $20$B$\to$$9$B) quantized to MXFP4 with a single Nemotron\,$+$\,SmolTalk mixture, owing to the high compute cost. We do not evaluate other families (Llama, Qwen, Mistral) or formats (NVFP4, INT4, FP8).

\textbf{Proprietary compression operator.} The operator producing the 60B and 9B students is proprietary. QAH is agnostic to the compression method---it needs only logit-level access to the uncompressed teacher---but we have not verified that the gains transfer to other structural-compression approaches such as layer pruning, SliceGPT, or low-rank decomposition.

\section{Conclusion}
\label{sec:conclusion}

We described Quantization-Aware Healing, the recipe we adopted for recovering LLMs that have been both structurally compressed and 4-bit quantized. QAH heals by distilling the compressed, quantized student from the original uncompressed model rather than from a recovered full-precision checkpoint. On a GPT-OSS 120B$\to$60B$\to$MXFP4 pipeline the resulting 4-bit model matches or beats its own bfloat16 source on 7 of 9 benchmarks and reaches the 120B model on LiveCodeBench, at roughly $4\times$ lower weight memory and half the parameter count. Against a matched QAT baseline it converges about $7\times$ faster and, unlike QAT, does not collapse under continued training---so it can be shipped without hand-tuned early stopping. Alongside the recipe we reported the operational lessons that made it work, including a large and reproducible quality gap between distributed-training backends, and we were explicit about the comparison we did not run. For practitioners, the practical message is that in a distillation-based healing pipeline the quantization step is not a cost to be minimized but an additional opportunity for teacher supervision, yielding a model that is simultaneously cheaper to serve, lighter in memory, and at least as accurate as its full-precision counterpart. The recipe is not merely a research prototype: it produced the open-weight Hypernova-60B release and a separate model built for an internal client deployment, both healed with the pipeline described here.


\bibliography{custom}

@article{liu2023llmqat,
  author    = {Zechun Liu and Barlas Oguz and Changsheng Zhao and Ernie Chang
               and Pierre Stock and Yashar Mehdad and Yangyang Shi and
               Raghuraman Krishnamoorthi and Vikas Chandra},
  title     = {{LLM-QAT}: Data-Free Quantization Aware Training for Large Language Models},
  journal   = {arXiv preprint arXiv:2305.17888},
  year      = {2023},
  url       = {https://arxiv.org/abs/2305.17888},
}

@inproceedings{frantar2023gptq,
  author    = {Elias Frantar and Saleh Ashkboos and Torsten Hoefler and Dan Alistarh},
  title     = {{GPTQ}: Accurate Post-Training Quantization for Generative Pre-trained Transformers},
  booktitle = {Proceedings of ICLR},
  year      = {2023},
  url       = {https://arxiv.org/abs/2210.17323},
}

@inproceedings{gu2024minillm,
  author    = {Yuxian Gu and Li Dong and Furu Wei and Minlie Huang},
  title     = {{MiniLLM}: Knowledge Distillation of Large Language Models},
  booktitle = {Proceedings of ICLR},
  year      = {2024},
  url       = {https://arxiv.org/abs/2306.08543},
}

@article{hinton2015distilling,
  author    = {Geoffrey Hinton and Oriol Vinyals and Jeff Dean},
  title     = {Distilling the Knowledge in a Neural Network},
  journal   = {arXiv preprint arXiv:1503.02531},
  year      = {2015},
  url       = {https://arxiv.org/abs/1503.02531},
}

@inproceedings{jacob2018quantization,
  author    = {Benoit Jacob and Skirmantas Kligys and Bo Chen and Menglong Zhu and
               Matthew Tang and Andrew Howard and Hartwig Adam and Dmitry Kalenichenko},
  title     = {Quantization and Training of Neural Networks for Efficient
               Integer-Arithmetic-Only Inference},
  booktitle = {Proceedings of CVPR},
  year      = {2018},
  url       = {https://arxiv.org/abs/1712.05877},
}

@inproceedings{jain2024livecodebench,
title={LiveCodeBench: Holistic and Contamination Free Evaluation of Large Language Models for Code},
author={Naman Jain and King Han and Alex Gu and Wen-Ding Li and Fanjia Yan and Tianjun Zhang and Sida Wang and Armando Solar-Lezama and Koushik Sen and Ion Stoica},
booktitle={The Thirteenth International Conference on Learning Representations},
year={2025},
url={https://openreview.net/forum?id=chfJJYC3iL}
}

@inproceedings{ko2024distillm,
  author    = {Jongwoo Ko and Sungnyun Kim and Tianyi Chen and Se-Young Yun},
  title     = {{DistiLLM}: Towards Streamlined Distillation for Large Language Models},
  booktitle = {Proceedings of ICML},
  year      = {2024},
  url       = {https://arxiv.org/abs/2402.03898},
}

@article{krishnamoorthi2018quantizing,
  author    = {Raghuraman Krishnamoorthi},
  title     = {Quantizing Deep Convolutional Networks for Efficient Inference:
               A Whitepaper},
  journal   = {arXiv preprint arXiv:1806.08342},
  year      = {2018},
  url       = {https://arxiv.org/abs/1806.08342},
}

@inproceedings{lin2024awq,
  author    = {Ji Lin and Jiaming Tang and Haotian Tang and Shang Yang and
               Wei-Ming Chen and Wei-Chen Wang and Guangxuan Xiao and Xingyu Dang
               and Chuang Gan and Song Han},
  title     = {{AWQ}: Activation-aware Weight Quantization for {LLM} Compression
               and Acceleration},
  booktitle = {Proceedings of MLSys},
  year      = {2024},
  url       = {https://arxiv.org/abs/2306.00978},
}

@article{micikevicius2022fp8,
  author    = {Paulius Micikevicius and Dusan Stosic and Neil Burgess and Marius Cornea
               and Pradeep Dubey and Richard Grisenthwaite and Sangwon Ha and
               Alexander Heinecke and Patrick Judd and John Kamalu and Naveen Mellempudi
               and Stuart Oberman and Mohammad Shoeybi and Michael Siu and Hao Wu},
  title     = {{FP8} Formats for Deep Learning},
  journal   = {arXiv preprint arXiv:2209.05433},
  year      = {2022},
  url       = {https://arxiv.org/abs/2209.05433},
}

@article{nvidia2026qad,
  author    = {{NVIDIA}},
  title     = {Quantization-Aware Distillation for {NVFP4} Inference Accuracy Recovery},
  journal   = {arXiv preprint arXiv:2601.20088},
  year      = {2026},
  url       = {https://arxiv.org/abs/2601.20088},
}

@article{openai2025gptoss,
  author    = {{OpenAI}},
  title     = {{GPT-OSS}: Open-weight Models Model Card},
  journal   = {arXiv preprint arXiv:2508.10925},
  year      = {2025},
  url       = {https://arxiv.org/abs/2508.10925},
}

@inproceedings{polino2018model,
  author    = {Antonio Polino and Razvan Pascanu and Dan Alistarh},
  title     = {Model Compression via Distillation and Quantization},
  booktitle = {Proceedings of ICLR},
  year      = {2018},
  url       = {https://arxiv.org/abs/1802.05668},
}

@inproceedings{rasley2020deepspeed,
  author    = {Jeff Rasley and Samyam Rajbhandari and Olatunji Ruwase and Yuxiong He},
  title     = {{DeepSpeed}: System Optimizations Enable Training Deep Learning Models
               with Over 100 Billion Parameters},
  booktitle = {Proceedings of KDD},
  year      = {2020},
}

@article{rein2023gpqa,
  author    = {David Rein and Betty Li Hou and Asa Cooper Stickland and
               Jackson Petty and Richard Yuanzhe Pang and Julien Dirani and
               Julian Michael and Samuel R.\ Bowman},
  title     = {{GPQA}: A Graduate-Level Google-Proof {Q\&A} Benchmark},
  journal   = {arXiv preprint arXiv:2311.12022},
  year      = {2023},
  url       = {https://arxiv.org/abs/2311.12022},
}

@article{rouhani2023microscaling,
  author    = {Bita Darvish Rouhani and Ritchie Zhao and Ankit More and
               Mathew Hall and Alireza Khodamoradi and Summer Deng and Dhruv Choudhary
               and Marius Cornea and Eric Dellinger and Kristof Denolf and
               Stosic Dusan and Venmugil Elango and Maximilian Golub and
               Alexander Heinecke and Phil James-Roxby and Dharmesh Jani and
               Gaurav Kaul and Chun-Chen Liu and Levi Melnick and Maral Mesmakhosroshahi
               and Andres Rodriguez and Michael Schulte and Rasoul Shafipour and
               Lei Shao and Michael Siu and Pradeep Dubey and Paulius Micikevicius
               and Maxim Naumov and Colin Verrilli and Ralph Wittig and
               Doug Burger and Eric Chung},
  title     = {Microscaling Data Formats for Deep Learning},
  journal   = {arXiv preprint arXiv:2310.10537},
  year      = {2023},
  url       = {https://arxiv.org/abs/2310.10537},
}

@inproceedings{wang2024mmlupro,
  author    = {Yubo Wang and Xueguang Ma and Ge Zhang and Yuansheng Ni and
               Abhranil Chandra and Shengding Hu and Aaran Zhu and Bo Li and
               Yiming Liang and Mingyu Ding and Beidi Chen and Jian Yang and
               Wenhu Chen},
  title     = {{MMLU-Pro}: A More Robust and Challenging Multi-Task Language
               Understanding Benchmark},
  booktitle = {Proceedings of NeurIPS},
  year      = {2024},
  url       = {https://arxiv.org/abs/2406.01574},
}

@inproceedings{xiao2023smoothquant,
  author    = {Guangxuan Xiao and Ji Lin and Mickael Seznec and Hao Wu and
               Julien Demouth and Song Han},
  title     = {{SmoothQuant}: Accurate and Efficient Post-Training Quantization
               for Large Language Models},
  booktitle = {Proceedings of ICML},
  year      = {2023},
  url       = {https://arxiv.org/abs/2211.10438},
}

@inproceedings{zhao2023fsdp,
  author    = {Yanli Zhao and Andrew Gu and Rohan Varma and Liang Luo and
               Chien-Chin Huang and Min Xu and Less Wright and Hamid Shojanazeri
               and Myle Ott and Sam Shleifer and Alban Desmaison and Can Balioglu
               and Pritam Damania and Bernard Nguyen and Geeta Chauhan and
               Yuchen Hao and Ajit Mathews and Shen Li},
  title     = {{PyTorch FSDP}: Experiences on Scaling Fully Sharded Data Parallel},
  booktitle = {Proceedings of VLDB},
  year      = {2023},
  url       = {https://arxiv.org/abs/2304.11277},
}

@inproceedings{ma2023llmpruner,
  author    = {Xinyin Ma and Gongfan Fang and Xinchao Wang},
  title     = {{LLM-Pruner}: On the Structural Pruning of Large Language Models},
  booktitle = {Proceedings of NeurIPS},
  year      = {2023},
  url       = {https://arxiv.org/abs/2305.11627},
}

@inproceedings{xia2022structured,
  author    = {Mengzhou Xia and Zexuan Zhong and Danqi Chen},
  title     = {Structured Pruning Learns Compact and Accurate Models},
  booktitle = {Proceedings of ACL},
  year      = {2022},
  url       = {https://arxiv.org/abs/2204.00408},
}

@inproceedings{ashkboos2024slicegpt,
  author    = {Saleh Ashkboos and Maximilian L. Croci and Marcelo Gennari do Nascimento
               and Torsten Hoefler and James Hensman},
  title     = {{SliceGPT}: Compress Large Language Models by Deleting Rows and Columns},
  booktitle = {Proceedings of ICLR},
  year      = {2024},
  url       = {https://arxiv.org/abs/2401.15024},
}

@article{wang2024svdllm,
  author    = {Xin Wang and Yu Zheng and Zhongwei Wan and Mi Zhang},
  title     = {{SVD-LLM}: Truncation-aware Singular Value Decomposition for
               Large Language Model Compression},
  journal   = {arXiv preprint arXiv:2403.07378},
  year      = {2024},
  url       = {https://arxiv.org/abs/2403.07378},
  note      = {Verify author list against arXiv before camera-ready.},
}

@article{yuan2023asvd,
  author    = {Zhihang Yuan and Yuzhang Shang and Yue Song and Qiang Wu and
               Yan Yan and Guangyu Sun},
  title     = {{ASVD}: Activation-aware Singular Value Decomposition for
               Compressing Large Language Models},
  journal   = {arXiv preprint arXiv:2312.05821},
  year      = {2023},
  url       = {https://arxiv.org/abs/2312.05821},
}

@inproceedings{novikov2015tensorizing,
  author    = {Alexander Novikov and Dmitry Podoprikhin and Anton Osokin
               and Dmitry Vetrov},
  title     = {Tensorizing Neural Networks},
  booktitle = {Proceedings of NeurIPS},
  year      = {2015},
  url       = {https://arxiv.org/abs/1509.06569},
}

@article{mcgowan2024fishleg,
  author    = {Jamie McGowan and Wei Sheng Lai and Weibin Chen and Henry Aldridge and
               Jools Clarke and Jezabel Garcia and Rui Xia and Yilei Liang and
               Guillaume Hennequin and Alberto Bernacchia},
  title     = {Efficient Model Compression Techniques with {FishLeg}},
  journal   = {arXiv preprint arXiv:2412.02328},
  year      = {2024},
  url       = {https://arxiv.org/abs/2412.02328},
}

@inproceedings{tomut2025compactifai,
  author    = {Andrei Tomut and Saeed S. Jahromi and Abhijoy Sarkar and Uygar Kurt and
               Sukhbinder Singh and Faysal Ishtiaq and Cesar Mu{\~n}oz and
               Prabdeep Singh Bajaj and Ali Elborady and Gianni {del Bimbo} and
               Mehrazin Alizadeh and David Montero and Pablo Martin-Ramiro and
               Muhammad Ibrahim and Oussama Tahiri Alaoui and John Malcolm and
               Samuel Mugel and Roman Orus},
  title     = {{CompactifAI}: Extreme Compression of Large Language Models using
               Quantum-Inspired Tensor Networks},
  booktitle = {Proceedings of the 33rd European Symposium on Artificial Neural
               Networks, Computational Intelligence and Machine Learning ({ESANN})},
  year      = {2025},
  url       = {https://arxiv.org/abs/2401.14109},
}

@article{yin2023outlier,
  author    = {Lu Yin and You Wu and Zhenyu Zhang and Cheng-Yu Hsieh and
               Yaqing Wang and Yiling Jia and Mykola Pechenizkiy and Yi Liang and
               Zhangyang Wang and Shiwei Liu},
  title     = {Outlier Weighed Layerwise Sparsity ({OWL}): A Missing Secret Sauce
               for Pruning {LLMs} to High Sparsity},
  journal   = {arXiv preprint arXiv:2310.05175},
  year      = {2023},
  url       = {https://arxiv.org/abs/2310.05175},
}

@article{xu2023tensorgpt,
  author    = {Mingxue Xu and Yao Lei Xu and Danilo P. Mandic},
  title     = {{TensorGPT}: Efficient Compression of the Embedding Layer in
               {LLMs} based on the Tensor-Train Decomposition},
  journal   = {arXiv preprint arXiv:2307.00526},
  year      = {2023},
  url       = {https://arxiv.org/abs/2307.00526},
}

@article{sanh2019distilbert,
  author    = {Victor Sanh and Lysandre Debut and Julien Chaumond and Thomas Wolf},
  title     = {{DistilBERT}, a Distilled Version of {BERT}: Smaller, Faster,
               Cheaper and Lighter},
  journal   = {arXiv preprint arXiv:1910.01108},
  year      = {2019},
  url       = {https://arxiv.org/abs/1910.01108},
}

@inproceedings{jiao2020tinybert,
  author    = {Xiaoqi Jiao and Yichun Yin and Lifeng Shang and Xin Jiang and
               Xiao Chen and Linlin Li and Fang Wang and Qun Liu},
  title     = {{TinyBERT}: Distilling {BERT} for Natural Language Understanding},
  booktitle = {Findings of EMNLP},
  year      = {2020},
  url       = {https://arxiv.org/abs/1909.10351},
}

@article{bondarenko2024llmqat,
  author    = {Yelysei Bondarenko and Riccardo Del Chiaro and Markus Nagel},
  title     = {Low-Rank Quantization-Aware Training for {LLMs}},
  journal   = {arXiv preprint arXiv:2406.06385},
  year      = {2024},
  url       = {https://arxiv.org/abs/2406.06385},
}

@inproceedings{wang2022efficientvlm,
  author    = {Tiannan Wang and Wangchunshu Zhou and Yan Zeng and Xinsong Zhang},
  title     = {{EfficientVLM}: Fast and Accurate Vision-Language Models via
               Knowledge Distillation and Modal-adaptive Pruning},
  booktitle = {Findings of ACL},
  year      = {2023},
  url       = {https://arxiv.org/abs/2210.07795},
}

@article{Nemotron_Cascade_2,
  title={Nemotron-Cascade 2: Post-Training LLMs with Cascade RL and Multi-Domain On-Policy Distillation},
  author={Yang, Zhuolin and Liu, Zihan and Chen, Yang and Dai, Wenliang and Wang, Boxin and Lin, Sheng-Chieh and Lee, Chankyu and Chen, Yangyi and Jiang, Dongfu and He, Jiafan and Pi, Renjie and Lam, Grace and Lee, Nayeon and Bukharin, Alexander and Shoeybi, Mohammad and Catanzaro, Bryan and Ping, Wei},
  year={2026}
}

@misc{nvidia_nemotron_nano_v3_2025,
  title  = {{Nemotron 3 Nano}: Open, Efficient Mixture-of-Experts Hybrid {Mamba}-{Transformer} Model for {Agentic} Reasoning},
  author = {{NVIDIA}},
  year   = {2025},
  url    = {https://arxiv.org/abs/2512.20848},
  note   = {Technical report}
}

@misc{aime25,
      title={American Invitational Mathematics Examination (AIME) 2025}, 
      author={Zhang, Yifan and Math-AI, Team},
      year={2025},
}

@article{pyatkin2026generalizing,
  title={Generalizing verifiable instruction following},
  author={Pyatkin, Valentina and Malik, Saumya and Graf, Victoria and Ivison, Hamish and Huang, Shengyi and Dasigi, Pradeep and Lambert, Nathan and Hajishirzi, Hanna},
  journal={Advances in Neural Information Processing Systems},
  volume={38},
  year={2026}
}

@article{tian2024scicode,
  title={Scicode: A research coding benchmark curated by scientists},
  author={Tian, Minyang and Gao, Luyu and Zhang, Shizhuo D and Chen, Xinan and Fan, Cunwei and Guo, Xuefei and Haas, Roland and Ji, Pan and Krongchon, Kittithat and Li, Yao and others},
  journal={Advances in Neural Information Processing Systems},
  volume={37},
  pages={30624--30650},
  year={2024}
}

@misc{aider,
  author       = {Gauthier, Paul},
  title        = {{Aider LLM Leaderboards: Polyglot Coding Benchmark}},
  howpublished = {\url{https://aider.chat/docs/leaderboards/\#polyglot-leaderboard}},
  year         = {2024},
  note         = {Accessed: 2024}
}

@dataset{artificialanalysis2025lcr,
  title={Artificial Analysis Long Context Reasoning Benchmark(LCR)},
  author={Artificial Analysis Team},
  year={2025},
  publisher={Artificial Analysis, Inc.}
}

@article{tomut2024compactifai,
  title={Compactifai: extreme compression of large language models using quantum-inspired tensor networks},
  author={Tomut, Andrei and Jahromi, Saeed S and Sarkar, Abhijoy and Kurt, Uygar and Singh, Sukhbinder and Ishtiaq, Faysal and Mu{\~n}oz, Cesar and Bajaj, Prabdeep Singh and Elborady, Ali and del Bimbo, Gianni and others},
  journal={arXiv preprint arXiv:2401.14109},
  pages={12},
  year={2024}
}

@article{bercovich2024puzzle,
  title={Puzzle: Distillation-based nas for inference-optimized llms},
  author={Bercovich, Akhiad and Ronen, Tomer and Abramovich, Talor and Ailon, Nir and Assaf, Nave and Dabbah, Mohammad and Galil, Ido and Geifman, Amnon and Geifman, Yonatan and Golan, Izhak and others},
  journal={arXiv preprint arXiv:2411.19146},
  year={2024}
}

@misc{qwen3.5,
    title  = {{Qwen3.5}: Towards Native Multimodal Agents},
    author = {{Qwen Team}},
    month  = {February},
    year   = {2026},
    url    = {https://qwen.ai/blog?id=qwen3.5}
}

@misc{bakouch2025smollm3,
  title={{SmolLM3: smol, multilingual, long-context reasoner}},
  author={Bakouch, Elie and Ben Allal, Loubna and Lozhkov, Anton and Tazi, Nouamane and Tunstall, Lewis and Patiño, Carlos Miguel and Beeching, Edward and Roucher, Aymeric and Reedi, Aksel Joonas and Gallouédec, Quentin and Rasul, Kashif and Habib, Nathan and Fourrier, Clémentine and Kydlicek, Hynek and Penedo, Guilherme and Larcher, Hugo and Morlon, Mathieu and Srivastav, Vaibhav and Lochner, Joshua and Nguyen, Xuan-Son and Raffel, Colin and von Werra, Leandro and Wolf, Thomas},
  year={2025},
  howpublished={\url{https://huggingface.co/blog/smollm3}}
}

@article{barres2025tau,
  title={tau2-Bench: Evaluating Conversational Agents in a Dual-Control Environment},
  author={Barres, Victor and Dong, Honghua and Ray, Soham and Si, Xujie and Narasimhan, Karthik},
  journal={arXiv preprint arXiv:2506.07982},
  year={2025}
}

@inproceedings{rajbhandari2020zero,
  title={Zero: Memory optimizations toward training trillion parameter models},
  author={Rajbhandari, Samyam and Rasley, Jeff and Ruwase, Olatunji and He, Yuxiong},
  booktitle={SC20: international conference for high performance computing, networking, storage and analysis},
  pages={1--16},
  year={2020},
  organization={IEEE}
}

@article{jansen2026blockremoval,
  title     = {Block Removal for Large Language Models through Constrained Binary Optimization},
  author    = {Jansen, David and Rausch, Roman and Hashemi, Ali and Montero, David and Or{\'u}s, Rom{\'a}n},
  journal   = {arXiv preprint arXiv:2602.00161},
  year      = {2026},
  url       = {https://arxiv.org/abs/2602.00161}
}

@misc{ryskulov2026chunkedkl,
  title         = {Offline Top-K Logits and a Fused Chunked {KL} Loss},
  author        = {Ryskulov, Bakbergen and Garc\'ia-Ferrero, Iker and Montero, David and Jansen, David and Hashemi, Ali and Garcia, Jezabel R. and Tiene, Antonio and Or\'us, Rom\'an},
  year          = {2026},
  eprint        = {2608.03796},
  archivePrefix = {arXiv},
  primaryClass  = {cs.LG},
  url           = {https://arxiv.org/abs/2608.03796}
}

\appendix
\section{QAT Training-Systems Sweep}
\label{app:sweep}

Table~\ref{tab:qat-sweep} reports the eleven-run QAT sweep on the 120B student that underlies the backend lesson in \S\ref{sec:lessons}. Every run starts from the same recovered checkpoint with the same frozen-submodule set, Nemotron\,$+$\,SmolTalk corpus, gradient clipping at $1.0$, and 100-step warmup. The variables probed are the distributed backend, effective sequence length, learning rate, batch size, cluster size, and layer-norm freezing strategy. Only the FSDP2 configuration (R11) reaches the released 120B MXFP4 baseline on GPQA Diamond; every DeepSpeed configuration plateaus $3.5$--$14.2$ points below it across a four-order-of-magnitude learning-rate range. R11 is the QAT arm used in the matched comparison in the main text.

\begin{table}[t]
  \centering
  \small
  \setlength{\tabcolsep}{3.5pt}
  \begin{tabular}{@{}llccccc@{}}
    \toprule
    \textbf{Run} & \textbf{Backend} & \textbf{SL} & \textbf{lr} & \textbf{GPQA:D} & \textbf{MMLU-P} & \textbf{LCB} \\
    \midrule
    R1  & FSDP2     & 2k  & $10^{-5}$          & $69.19$ & $70.86$ & $61.94$ \\
    R2  & DeepSpeed & 4k  & $10^{-5}$          & $65.15$ & $69.15$ & $66.42$ \\
    R3  & FSDP2     & 2k  & $10^{-5}$          & $65.66$ & $70.28$ & $64.93$ \\
    R4  & DeepSpeed & 4k  & $10^{-5}$          & $65.15$ & $69.39$ & $63.81$ \\
    R5  & DeepSpeed & 8k  & $5{\times}10^{-6}$ & $65.15$ & $69.45$ & $66.79$ \\
    R6  & DeepSpeed & 8k  & $5{\times}10^{-6}$ & $63.64$ & $68.89$ & $63.43$ \\
    R7  & DeepSpeed & 8k  & $5{\times}10^{-4}$ & $59.09$ & $66.86$ & $56.72$ \\
    R8  & DeepSpeed & 8k$_f$ & $10^{-7}$       & $65.15$ & $69.22$ & $67.54$ \\
    R9  & DeepSpeed & 8k$_f$ & $5{\times}10^{-7}$ & $58.59$ & $69.11$ & $64.18$ \\
    R10 & DeepSpeed & 8k$_f$ & $10^{-6}$       & $64.14$ & $69.25$ & $65.30$ \\
    \textbf{R11} & \textbf{FSDP2} & \textbf{2k} & $\mathbf{10^{-5}}$ & $\mathbf{73.74}$ & $\mathbf{70.95}$ & $\mathbf{66.42}$ \\
    \midrule
    \multicolumn{4}{@{}l}{\textit{Target (120B MXFP4)}} & $72.73$ & $79.47$ & $68.66$ \\
    \bottomrule
  \end{tabular}
  \caption{QAT sweep (R1--R11) over the 120B student; best checkpoint score per metric. R11 repeats R1 with $\sim$$2\times$ more steps. 8k$_f$ denotes 8k context with a position-id fix. R11 is the sole run reaching the MXFP4 baseline on GPQA Diamond and is the QAT reference arm in the main text.}
  \label{tab:qat-sweep}
\end{table}

\end{document}